\documentclass[preprint,11pt, 3p]{elsarticle}

\usepackage{amssymb}

\usepackage{subcaption}
\usepackage{natbib}

\usepackage{tabulary}
\usepackage{multirow}
\usepackage{array}

\usepackage{mathtools}

\newcolumntype{P}[1]{>{\centering\arraybackslash}p{#1}}
\newcolumntype{M}[1]{>{\centering\arraybackslash}m{#1}}

\usepackage[ruled,linesnumbered]{algorithm2e}
\usepackage[noend]{algpseudocode}

\begin{document}

\begin{frontmatter}



\title{Financial Trading Model with Stock Bar Chart Image Time Series with Deep Convolutional Neural Networks}


\author{Omer Berat Sezer, Ahmet Murat Ozbayoglu}
\address{TOBB University of Economics and Technology}
\address{Department  of Computer Engineering}
\address{Ankara, 06560, Turkey}

\begin{abstract}
Even though computational intelligence techniques have been extensively utilized in financial trading systems, almost all developed models use the time series data for price prediction or identifying buy-sell points. However, in this study we decided to use 2-D stock bar chart images directly without introducing any additional time series associated with the underlying stock. We propose a novel algorithmic trading model CNN-BI (Convolutional Neural Network with Bar Images) using a 2-D Convolutional Neural Network. We generated 2-D images of sliding windows of 30-day bar charts for Dow 30 stocks and trained a deep Convolutional Neural Network (CNN) model for our algorithmic trading model. We tested our model separately between 2007-2012 and 2012-2017 for representing different market conditions. The results indicate that the model was able to outperform Buy and Hold strategy, especially in trendless or bear markets. Since this is a preliminary study and probably one of the first attempts using such an unconventional approach, there is always potential for improvement. Overall, the results are promising and the model might be integrated as part of an ensemble trading model combined with different strategies.
\end{abstract}


\end{frontmatter}


\section{Introduction}
\label{Introduction}
In recent years, algorithmic models in finance using computational intelligence and machine learning techniques became very popular. Financial time series data is analyzed, prediction techniques are adapted and fast algorithmic trading methods are implemented using various computational models for stock and forex markets \citep{Cavalcante2016}.

Machine learning algorithms are used in financial analysis especially in price prediction of stocks, Exchange-Traded Funds (ETFs), options. Support vector machines, artificial neural networks, evolutionary algorithms, hybrid and ensemble methods are among the mostly preferred machine learning algorithms. With the introduction of deep learning models, methods such as recurrent neural network (RNN) \citep{Krauss2017}, convolutional neural network (CNN) \citep{Chen2016}, and long short term memory (LSTM) \citep{Fischer2017} have started appearing on implementations that are used for analyzing financial data. However, developed deep learning models for financial analysis are still limited in number.

CNN is the most common deep learning method that is used for classification of two-dimensional images. The implementations for two-dimensional image classification has increased with improving success in recent years. As one of the early proposed models, AlexNet achieved ~50-55\% success rate. Later, GoogleNet, ENet, ResNet-18, VGG were proposed, their success rates were approximately ~65-70\%.  Following these improvements, different versions of Inception (v3, v4) and ResNet (v50, v101, v152) algorithms were proposed for image classification and their success rates were approximately ~75-80\% \citep{Canziani2016}.

In this study, a novel approach that converts one-dimensional financial time series into two-dimensional images for an algorithmic trading model is proposed (Convolutional Neural Network with Bar Images: CNN-BI). With this model, time series data is converted to series of images that consist of bar chart representations of stock prices and each image is labeled with "Buy", "Sell" and "Hold". Each image contains 30-days of stock price data resulting in a 30x30 pixels image. Using deep convolutional neural network structure, the developed model is trained. Finally, predicted results are fed into a trading model and evaluated using real financial out-of-sample test data. To best of our knowledge, such an approach using bar chart images with deep CNN training and integrating it into a buy-sell decision support system is not studied in the literature previously.

The rest of the paper is structured as follows. After this introduction, the related work is provided in Section~\ref{RelatedWork} Related Work. Proposed method, proposed algorithm, formulations and implementation of this method are explained in Section~\ref{Method} Method. Financial evaluation of the proposed model and test results are analyzed and evaluated in Section~\ref{Evaluation} Evaluation. Lastly, conclusion is presented in Section~\ref{Conclusion} Conclusion.

\section{Related Work}
\label{RelatedWork}

Forecasting the trend and future price of an asset class is an important problem in financial systems. There are mainly two approaches that investors prefer to adapt for analyzing the asset prices: Fundamental analysis and technical analysis. Fundamental analysis can be implemented by getting detailed information about the financial statements of the underlying asset (mostly stock or exchange traded fund - ETF) such as cash flows, growth rate, discount factor, discount per year, capitalization rate, value at the end of year five, and present value of residual value. Fundamental analysis is generally used for deciding which stock should be chosen for trading. Whereas, the other common financial analysis approach, technical analysis is mainly concerned about the price movements of the underlying asset, hence generally the focus is on time series and mathematical analysis techniques. In literature, several different technical analysis indicators are used: Relative strength index (RSI), Williams \%R, simple moving average (SMA), exponential moving average (EMA), commodity channel index (CCI), moving average convergence and divergence (MACD), percentage price oscillator (PPO), rate of change (ROC), directional movement indicator (DMI), and parabolic SAR. In addition, new technical indicators are still developed and researched. 

\begin{figure}[ht]
\centering
\includegraphics[width=3in]{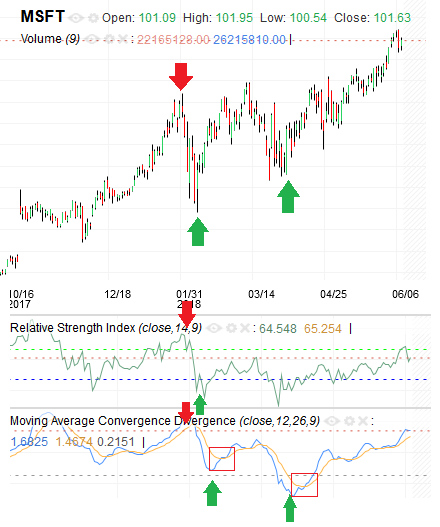}
\caption{Stock Trading Example with RSI and MACD}
\label{fig_stockTrading}
\end{figure}

Successful asset/stock trading is the most important issue for investors. Profitable stock trading has different points to be considered: selecting the appropriate stock for trading and timing of the buy-sell points. Choosing stocks in the market is the starting point for a successful trading system. Selected stock should have a good fundamental analysis background (cash flows, growth rate, financial statement, etc.). In addition, it is better to select a stock with high liquidity to avoid manipulation and speculation. After choosing the stock, the buy and sell timing needs to be decided. Investors mostly use technical analysis to detect best buy and sell points. An example of usage of technical indicators in stock trading is illustrated in Figure~\ref{fig_stockTrading}. In the example scenario, MSFT (Microsoft Corp.) stock is selected for trading, RSI and MACD are selected as the decisive technical indicators.  The most common usage of RSI and its interpretation is as follows: If the RSI value is over 70, the stock is considered to be in the “overbought” region indicating a sell signal. If the RSI value is under 30, the stock is assumed to be in the “oversold” region indicating a buy signal.  As illustrated in Figure~\ref{fig_stockTrading} green and red arrows on the RSI graph is point of the buy and sell, respectively. Meanwhile, MACD common usage and interpretation is as follows: If MACD line crosses signal lines in upward direction, it is predicted that stock prices will increase. In contrast, if MACD line crosses signal lines in downward direction, it is interpreted that stock prices will decrease.  The green and red arrows indicating buy and sell points are illustrated in Figure~\ref{fig_stockTrading}, for MACD. If the stock is bought and sold according to the technical analysis indicators, trading can be profitable. Signal points are optimized with different usage and combination of the technical analysis indicators. In our proposed approach, we try to find best buy and sell points using only stock prices.

With the development of computational intelligence and increasing computational capabilities, algorithmic trading models became popular in the recent years. Different strategies are adapted for the algorithmic trading models ranging from the ones that use raw time series data to models that depend on technical analysis or other triggers so that assets can be sold and bought automatically with the associated algorithms. Moreover, nowadays, researchers and developers also study new methods and models to predict financial time-series data by using artificial intelligence, machine learning, evolutionary algorithms and deep learning.

\subsection{Machine Learning and Evolutionary Solutions}

Machine learning (ML) algorithms are generally divided into three main groups: Supervised, unsupervised and reinforcement learning. In supervised learning, labeled data is used for training and test phases. There are different supervised learning techniques and models. Most common supervised learning techniques are listed as follows: Artificial neural networks, support vector machines, Bayesian networks, decision trees, ensembles of classifiers and instance-based learning models. In unsupervised learning, data is clustered and grouped with associated features. In this particular case, it does not make a difference if the data is labeled or not. Self-organizing maps, associated rule learning and clustering techniques are among the preferred unsupervised learning models. Reinforcement learning (RL) is another type of machine learning technique such that it uses the reinforcement signal to train the system. In RL, actions, states, transitions, rewards and discounts values are determined to implement learning mechanism. Value-based, policy-based and model (actor)-based approaches are the main choices for RL systems. Q-learning is the most common technique in ML solutions.  

Machine learning techniques are used in different areas such as image/video processing, classification and recognition processes, natural language processing, and time series data analysis. ML approaches can also be used to analyze financial time-series data \citep{Wang2011, Liao2010, Kašćelan2015, Enke2013}. Cavalcante et al. \citep{Cavalcante2016} reviewed all possible prediction model techniques that are used in financial time-series data. Support vector machines, artificial neural networks, hybrid mechanisms, optimization and ensemble methods are among the methods that are surveyed. Kašćelan et al. \citep{Kašćelan2015} developed a method that uses support vector machine and decision tree models to extract rules for stock market decision-makers. Chen et al. \citep{Chen2003} developed a method that uses artificial neural network model to predict the Taiwan Stock Index. Guresen et al. \citep{Guresen2011} used dynamic artificial neural networks and multi-layer perceptron (MLP) to forecast NASDAQ stock index. 

Sezer et al. \citep{Sezer2017_Artif} proposed an artificial neural network that uses the technical analysis indicators such as MACD, RSI, Williams \%R to predict Dow Jones 30 Stocks. Besides these approaches, evolutionary and genetic algorithms are used to forecast stock market prices and index in literature \citep{AguilarRivera2015}. Krollner et al.\citep{Krollner2010} surveyed stock market forecasting papers that use artificial neural networks, evolutionary optimization techniques, and hybrid methods.

\subsection{Deep Learning Solutions}

Deep neural networks that utilize deep learning are particular types of Artificial Neural Networks that consist of multiple layers \citep{LeCun2015}. There are different kinds of deep learning models. Convolutional Neural Network (CNN), Recurrent Neural Network (RNN), Deep Belief Networks, Restricted Boltzmann Machines (RBMs), and Long Short Term Memory (LSTM) networks are popular choices for deep learning models.

Backpropagation with multiple hidden layers was the first deep learning network to process data using stochastic gradient descent to update weights between layers. This type of network is no different than a fully-connected feedforward neural network, however, the number of hidden layers are significantly higher than the ones existing in traditional machine learning models. According to \citep{LeCun2015}, there are three major types of feedforward deep learning networks: fully-supervised; all layers unsupervised except the last one, which is supervised (but trained all together); and finally, all layers unsupervised except the last one, however, only the last supervised layer is trained explicitly. The first model is chosen when there is massive labeled training data available. The second model is generally preferred when only a limited number of labeled data are available (however, the overall data set can still be huge). The last model might be more suitable for cases when there is missing and/or mislabeled data.

Deep RNN is a type of deep learning network that is used for time series, sequential data, such as language and speech. RNNs are also used in traditional machine learning models (backpropagation through time, Jordan-Elman networks, etc.), however the time lengths in such models are generally less than the models used in deep RNN models. Deep RNNs are preferred due to their ability to include longer time periods. One standout deep RNN model is the LSTM network, where the network can remember both short term and long term values. LSTM networks are the preferred choice of many deep learning model developers when tackling complex problems like automatic speech recognition, and handwritten character recognition.

Meanwhile, deep learning networks are still the most frequently adapted choices for vision or image processing based classification problems, and CNN is the most common model adapted in these problems. CNN is mostly used in image and video processing, classification and recognition processes \citep{Krizhevsky2012, Karpathy2014, Lawrence1997, Ciresan2011}, sentence classification, and natural language process \citep{Kim2014, Kalchbrenner2014}.  CNN consists of convolutional and pooling layers that are applied on two-dimensional and visual data for processing, and gives a better solution when analyzing visual data.

In addition, there are different solutions that use deep learning methods to analyze financial time-series data in literature. Ding et al. \citep{Ding2015} proposed a deep learning model to predict stock prices by extracting texts from newspapers and internet. In their study, a deep convolutional neural network and neural tensor network are used to model short-term and long-term effects on stock price fluctuations.  S\&P500 stock data is used in their study. Sezer et al. \citep{Sezer2018} proposed a novel algorithmic trading model CNN-TA (Convolutional Neural Network with Technical Analysis) using a 2-D convolutional neural network and technical analysis. 15 different technical analysis indicators with 15 days period are used and 15x15 pixel images are constructed. Performance of the model is promising. 

Langkvist et al. \citep{Langkvist2014} reviewed the deep learning models (Boltzmann machine, autoencoder, recurrent neural network, convolution and pooling, and hidden Markov model) which are used for the analysis of time-series data, especially stock market indexes and stock prices. Besides, Sezer et al. \citep{Sezer2017_Deep} used deep feedforward neural network with optimized technical analysis parameters that are selected by using evolutionary algorithms (genetic algorithm). The result of the study shows that deep learning with evolutionary algorithms can be successful for determining buy-sell points for individual stocks over long out-of-sample test periods. Kwon et al. \citep{Kwon2007} proposed a recurrent neural network solution with genetic algorithms to predict stock values. Stock prices in NYSE and NASDAQ from 1992 to 2004 are used in their work. Fischer et al. \citep{Fischer2017} used LSTM to predict out-of-sample directional movements for stocks of the S\&P500 from 1992 until 2015. With LSTM technique, they have observed the LSTM method outperforming memory-free classification methods like random forest, deep neural nets, and logistic regression classifier. Krauss et al. \citep{Krauss2017} compared deep neural nets, gradient-boosted-trees, random forests, several ensembles of these methods to predict stock prices in S\&P500 from 1992 until 2015.

\section{Method}
\label{Method}

\begin{figure*}[ht]
\centering
\includegraphics[width=5in]{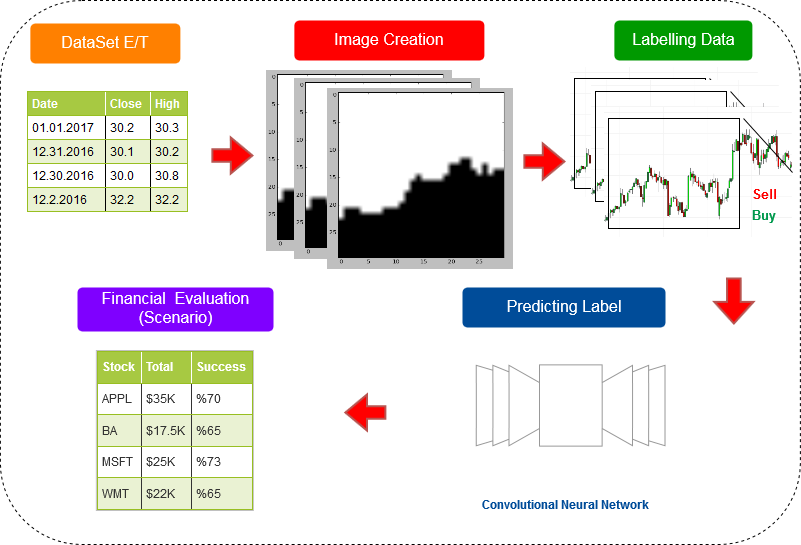}
\caption{Proposed Method }
\label{fig_phaseMethod}
\end{figure*}

In this study, an unconventional approach for stock forecasting is proposed that uses convolutional neural network to determine "Buy", "Sell" and "Hold" scenarios through images constructed from stock charts. In this model, time series data is converted into series of images that consist of bar charts of stock prices. Each converted image contains stock prices (y-axis) and time (x-axis). Also, each image contains information representing 30-days of stock prices that provide 30x30 (width and height of each converted images) pixels image. As can be seen in Figure~\ref{fig_phaseMethod}, our proposed method is divided into five main phases: dataset extract/transform, image construction, labeling each image, predicting label (convolutional neural network  analysis) and financial evaluation. Our aim is to find the most suitable points within the time series of the associated stock prices for buy-sell transactions to maximize the profit of algorithmic trading.

In image creation phase, time series stock values are used for constructing a binary image obtained from a normalized bar graph. In the first step, stock prices are normalized by passing the 30-day window over the daily close prices of the corresponding stock. Each daily stock value is illustrated as a price bar chart. 30-day values are combined to create a 30x30 time series image.  In our study, there are approximately 2500 images (1/1/1997 to 12/31/2006)  and 3750 images (1/1/1997 to 12/31/2012) for each stock price training data and approximately 2500 images (1/1/2007 to 1/1/2017) and 1250 images (1/1/2012 to 12/31/2017) for each stock price test data. For each stock, different training and test data image files are prepared and  evaluated seperately for their different characteristic features.

\begin{figure*}[ht]
\centering
\includegraphics[width=5in]{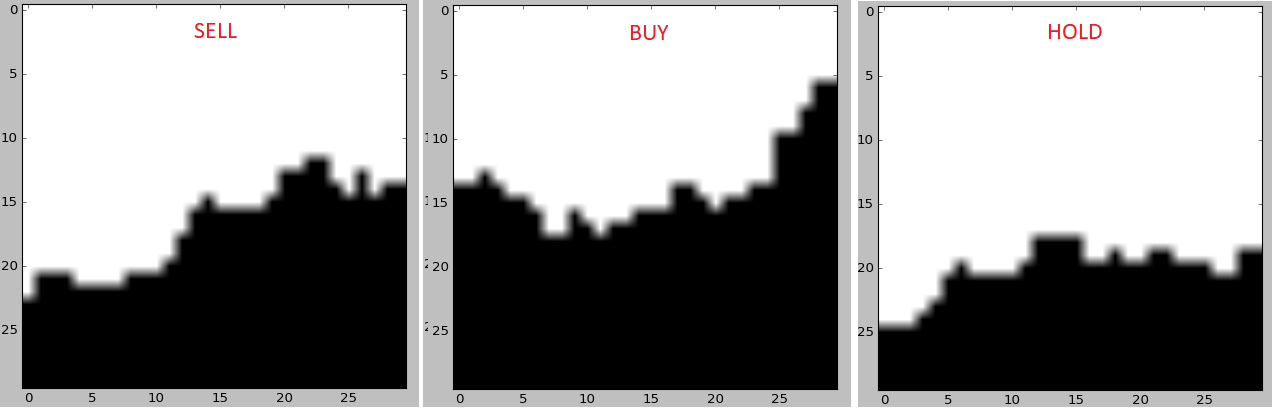}
\caption{Sample of 30x30 Pixels Images and Their Labels }
\label{fig_processLabel}
\end{figure*}

In labeling data phase, each image is labeled in order to represent the future trend of the prices. As illustrated in Equation~\ref{eq:slopeRef}, each reference slopes are calculated and stored in the list for each image. While labeling each image as "Buy","Hold", and "Sell", current slope of each image is also calculated (Equation~\ref{eq:slopeCurrent}). "TrendLabel" is defined for each image as shown in Equation~\ref{eq:trendLabel} (l denotes "TrendLabel"). After labeling the data, each label and its associated image are combined in a file for the learning phase. The aim of this method is to find whether buy-sell decisions can be made solely by using labeled bar chart images as inputs to a convolutional neural network and training the model accordingly. Figure~\ref{fig_processLabel} illustrates the sample images and their labels.

\begin{equation}\label{eq:slopeRef} 
slopeRef[1..n] =\frac{value34 - value30}{day34 - day30}
\end{equation}

\begin{equation}\label{eq:slopeCurrent}  
slopeCurrent = \frac{value45 - value30}{day45 - day30} 
\end{equation}

\begin{equation}\label{eq:trendLabel} 
l =
\begin{cases}
1=\text{('Buy')} , & \text{if slopeCurrent }>\text{slopeRef}[3n/5] \\
0=\text{('Hold')}, & \text{if (slopeRef}[2n/5]<\text{slopeCurrent)}\\
  & \text{and (slopeRef}[3n/5]>\text{slopeCurrent)}\\
2=\text{('Sell')}, & \text{if slopeCurrent }<\text{slopeRef}[2n/5] 
\end{cases}
\end{equation}

During labeling phase, the images are labeled by comparing the instantaneous slope (far future slope) with the average slope (near future slope) in the images. The processes during the labeling phase can also be implemented using different methods. The purpose of labeling is to benefit from the comparison of the average near-future slope and far-future slope values for CNN model implementation. The current model design is proposed for proof of concept purposes. The performance can be improved by suggesting different labeling mechanisms for future work.

Each of the images created in the model consists of 30x30 pixels. While the stock price values between days 0 and 30 are shown as bar chart, future values  (34th day price value = next image, 4th day price value; 45th day price value = next image, 15th day price value) are used for slope calculations during labeling. 

\begin{figure}[ht]
\centering
\includegraphics[width=3.5in]{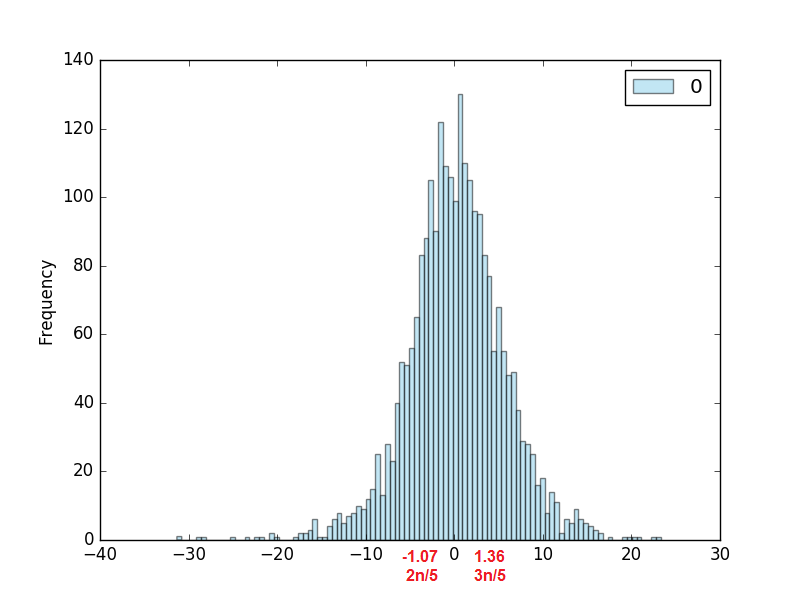}
\caption{An Example of Gaussian Distribution of the Slope Reference List and Seperation Values}
\label{fig_INTC_histogram}
\end{figure}

The slope between the price value on the 34th day and the price value on the 30th day is calculated to find the average slope reference list for each image in the training data (slopeRef = slopeReferenceList). All reference slopes are sorted and stored in a list before the labeling phase ([1..n] indicates a list that contains n number of slope data). The histogram of the reference slope values in the training data have a gaussian distribution. In order to use the slope values in the training data as references, the slope values at certain points are compared with the instant slope values (slopeCurrent). The separation points of the reference slope values in the list are calculated experimentally.  Figure~\ref{fig_INTC_histogram} shows an example of gaussian distribution of the slope reference list and separation values (INTC stock prices in 1997-2007 are used for reference slope histogram illustration)

The slope between the price value on the 45th day and the price value on the 30th day is calculated to find current slope (slopeCurrent) for the each image in the training dataset. The calculated instantaneous slope  (slopeCurrent) compared with the points on the reference slope list for labeling of each image ('Buy', 'Sell', 'Hold').

\begin{figure}[ht]
\centering
\includegraphics[width=6in]{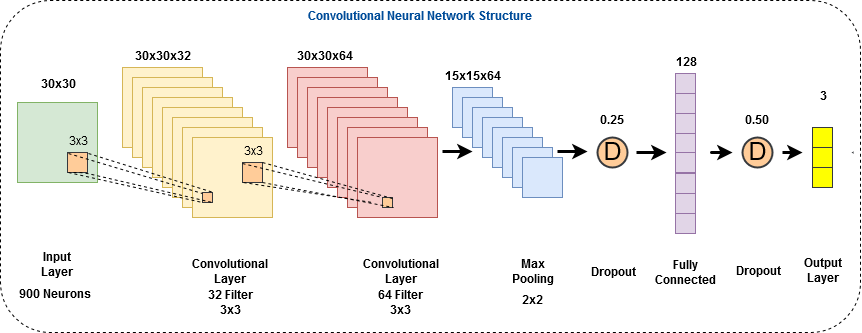}
\caption{CNN Process}
\label{fig_processCNN}
\end{figure}

In the learning phase, a convolutional neural network (CNN) is constructed to train and test the aforementioned data. Constructed CNN (Figure~\ref{fig_processCNN}) consists of eight layers in terms of  input layer (30x30), two convolutional layers (30x30x32, 30x30x64), a max pooling layer (15x15x64), two dropout layers (0.25, 0.50), fully connected layer (128), and an output layer(3). Hyper parameters are fine tuned through observing results from different experiments. Dropout values are tuned with different values: 0.1, 0,25, 0.5. Filter size is optimized with chosen of 3x3 filter size. In literature, different sizes of CNN filters are adapted: 3x3, 5x5 and 7x7. Decreasing filter size generally results in catching more details of the images. With the usage of 3x3 filter size, closest neighbors' information is used as input for the convolution operation (right, left, upper, lower, lower left, lower right, upper left, upper right). In our proposal, we preferred 3x3 filter size, because of  relatively small images (30x30 pixels) used as input. Hence,  significant intensity variations within the images can be noticed. In addition, CNN structure in the proposed model is similar to the deep CNN used in the MNIST algorithm. LeNet CNN structures, \citep{LeCun1995}  the deep CNN model with first successful results, consist of six layers. Adding more layers in CNN structure increases the complexity of the model. Without using a large training dataset, such a complex structured network might cause overfit and  reduce the accuracy on the test data. In future work, deeper structures can be used with larger training and test data. With the proposed model, the proof of concept design is implemented and tested. 

In Convolutional Neural Network structure, Convolutional layers have convolution operation.  Equation~\ref{eq:convolution} illustrates the convolution operation in one axis (t denotes time). The working structure of the convolution operation in CNN requires the usage of two dimensional images. Equation~\ref{eq:convolution2d} shows the convolution operation of a two dimensional image (I denotes input image, K denotes the kernel). In addition, Equation~\ref{eq:cnn} explains the neural network architecture (W denotes weights, x denotes input and b denotes bias). In the final stage of the network, softmax function is used to get the output. Equation~\ref{eq:softmax} illustrates the softmax function (y denotes output) \citep{Goodfellow2016}.

\begin{equation}\label{eq:convolution} 
s(t) = (x*w)(t) = \sum\limits_{a=-\infty}^\infty x(a)w(t-a) 
\end{equation}

\begin{equation}\label{eq:convolution2d} 
S(i,j) =  (I * K)(i, j) = \sum\limits_{m}\sum\limits_{n} I(m, n)K(i-m, j-n).
\end{equation}

\begin{equation}\label{eq:cnn} 
e_i = \sum\limits_{j} W_i,_j x_j +b_i.
\end{equation}

\begin{equation}\label{eq:softmax} 
y = softmax(e)
\end{equation}

\begin{algorithm*}[ht]
\centering
\fontsize{9}{11}\selectfont
\caption{Proposed Model Procedure}
\label{algo:algorithmCNN}
\begin{algorithmic}[1]
\Procedure{AllPhases()}{}
\State \textbf{Phase~DataSet~E/T:}
\State $dataset = read (open, close, high, low, adjustedClose, volume)$
\State \textbf{Phase~Image~Creation:}
\State $create~30x30~pixels~images$
\State $trainingDataset1 = dataset.split(dates=1997-2006)$
\State $trainingDataset2 = dataset.split(dates=1997-2012)$
\State $testDataset1= dataset.split(dates=2007-2012)$
\State $testDataset2 = dataset.split(dates=2012-2017)$
\State \textbf{Phase~Labelling~Data:}
\State $slopeRef[1..n]=calculateSlopReferences(farFutureValue=34, nearFutureValue=30)$
\State $calculate~distribution~of~the~class(Buy, Sell, Hold)~to~find~seperation~values$
\State $firstSepPoint, secondSepPoint=find~the~seperation~values(slopeRef[1..n])$
\State $for(all~images):$
\State $~~~~slopeCurrent=calculateEachImageSlope(farFutureValue=45, nearFutureValue=30)$
\State $~~~~if(slopeCurrent>=secondSepPoint):$
\State $~~~~~~~label=1~("Buy")$
\State $~~~~elif(slopeCurrent>firstSepPoint~and~slopeCurrent<secondSepPoint ):$
\State $~~~~~~~label=0~("Hold")$
\State $~~~~elif(slopeCurrent<=secondSepPoint):$
\State $~~~~~~~label=2~("Sell")$
\State $~~~~merge~labels~and~images$
\State $create~images~file$
\State \textbf{Phase~Predicting~Label:}
\State $trainingDataset1 = resample(trainingDataset1)~to~solve~data~imbalance~problem$
\State $trainingDataset2 = resample(trainingDataset2)~to~solve~data~imbalance~problem$
\State $model1 = CNN(epochs=100, blocksize=1028)$
\State $model1.train(trainingDataset1)$
\State $model1.test(testDataset1)$
\State $model2 = CNN(epochs=100, blocksize=1028)$
\State $model2.train(trainingDataset2)$
\State $model2.test(testDataset2)$
\State \textbf{Phase~Financial~Evaluation:}
\State $financialEvaluationScenario()$
\State $calculateEvaluationConstraints()$
\EndProcedure
\end{algorithmic}
\end{algorithm*}

CNN phase is implemented by using Keras\footnote{https://keras.io/}, Tensorflow\footnote{https://github.com/tensorflow} infrastructure and each test run lasts for 100 epochs. Algorithm~\ref{algo:algorithmCNN} summarizes the algorithm of the proposed method.

\section{Evaluation }
\label{Evaluation}

In the last phase, the trading model performance through the buy-sell transaction results are evaluated using the financial evaluation method. Each stock is bought, sold or held according to the predicted label. Financial evaluation scenario is illustrated in Equation~\ref{eq:financialEvaluation} ("S" denotes financial evaluation scenario, "tMoney" denotes totalMoney, \#OfStocks denotes "numberOfStocks").

\begin{equation}\label{eq:financialEvaluation} 
S=
\begin{cases}
\#OfStocks= \frac{tMoney}{price}, & \text{if label='Buy' } \\
no~action, &  \text{if label='Hold' }\\
tMoney=price * \#OfStocks  & \text{if label='Sell' }
\end{cases}
\end{equation}

If the predicted label related stock price is ``Buy", stock is bought at that point with all of the current available capital. If the label is ``Sell", stock is sold at that price. If the predicted label is ``Hold", there is no action performed at that point. Meanwhile, through this scenario, if the same label comes consecutively, the action is taken only with the first label and the corresponding transaction is performed. Repeating labels are ignored until the associated label changes. Starting capital for financial evaluation is \$10,000.00, trading commission is \$1.00 per transaction.

The proposed method is evaluated with Dow Jones 30 Stocks with different time periods (2007-2012 and 2012-2017).  The following evaluation metrics are adapted throughout the study. (Table~\ref{table:dow30_all} and Table~\ref{table:average} ): Our proposed CNN strategy annualized return (CNN-BI) ("AR" denotes annualized return) (Equation~\ref{eq:annualizedReturn}), "Buy and Hold" annualized return (BaH) (Equation~\ref{eq:annualizedReturn}), annualized number of transaction (AnT) (Equation~\ref{eq:annualizedNumberofTransaction}), percent of success (PoS) (Equation~\ref{eq:pos}), average percent profit per transactions (ApT) (Equation~\ref{eq:appt}), average transaction length (L) (Equation~\ref{eq:atl}), maximum profit percentage in transaction (MpT), maximum loss percentage in transaction (MlT), maximum capital (MaxC), minimum capital (MinC), idle ratio (IdleR) (Equation~\ref{eq:idleRatio}), Average Percentage of Maximum Drawdown (MDD), Return Over Maximum Drawdown (RoMaD). Formulas for the corresponding evaluation metrics are illustrated in Equations~\ref{eq:annualizedReturn}, \ref{eq:annualizedNumberofTransaction}, \ref{eq:pos}, \ref{eq:appt}, \ref{eq:atl}, \ref{eq:idleRatio}.

\begin{equation}\label{eq:annualizedReturn}
AR= ({(\frac{totalMoney}{startMoney})^{\frac{1}{numberOfYears}}}-1)*100
\end{equation}

\begin{equation}\label{eq:annualizedNumberofTransaction} 
AnT = \frac{transactionCount}{numberOfYears}
\end{equation}

\begin{equation}\label{eq:pos} 
PoS = \frac{successTransactionCount}{transactionCount}*100
\end{equation}

\begin{equation}\label{eq:appt} 
ApT = \frac{totalPercentProfit}{transactionCount}*100
\end{equation}

\begin{equation}\label{eq:atl} 
L = \frac{totalTransactionLength}{transactionCount}*100
\end{equation}

\begin{equation}\label{eq:idleRatio} 
IdleR = \frac{data.length-totalTransLength}{data.length}*100
\end{equation}

\begin{figure*}[ht]
\centering
\includegraphics[width=6in]{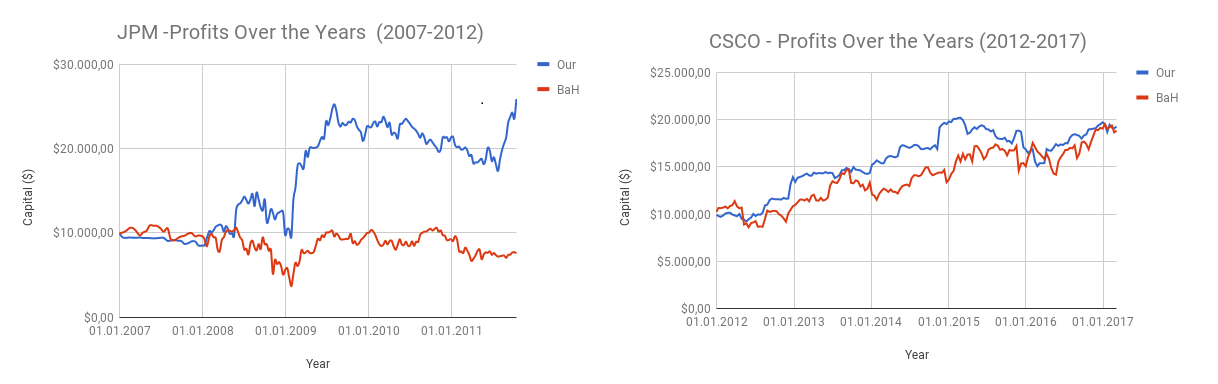}
\caption{Comparison of the Proposed Algorithm and BaH Method Results}
\label{fig_graphStocks}
\end{figure*}

Two different training and testing periods were chosen to represent different market conditions. In the first case, the model was trained using the data between 1997-2007 and the test period was between 2007-2012 which included the 2008 financial crisis. During that particular time span, the stock market made violent swings resulting in a very highly volatile period. In the second case, the model was trained with 1997-2012 data while the period between 2012-2017 was adapted for the out-of sample test period which was mainly consisted of a steady bull market with relatively low volatility. In each case the model was not retrained with trailing data just to see if the trained model was able to perform successfully over a long period of time to test the reliability and robustness of the trained model. Figure~\ref{fig_graphStocks} illustrates the comparison of the proposed algorithm  and BaH results. Table~\ref{table:transactionSample} shows the sample of the transactions of JPM.

\begin{table}[]
\fontsize{8}{10}\selectfont
\centering
\caption{Transactions Sample: Transactions OF JPM}
\label{table:transactionSample}
\begin{tabular}{ l l l l }
\hline
\textbf{Transaction Number} & \textbf{Interval (Day)} & \textbf{Gain} & \textbf{Instant Capital} \\ \hline
\textbf{1}                  & 21-25                   & -\$561,19     & \$9.481,81               \\ \hline
\textbf{16}                 & 117-117                 & \$60,62       & \$9.120,70               \\ \hline
\textbf{40}                 & 300-301                 & \$415,05      & \$10.747,49              \\ \hline
\textbf{60}                 & 402-403                 & \$1.666,47    & \$14.833,34              \\ \hline
\textbf{80}                 & 537-543                 & \$2.545,37    & \$17.889,30              \\ \hline
\textbf{100}                & 680-683                 & \$1.385,12    & \$23.095,23              \\ \hline
\textbf{120}                & 806-807                 & \$558,66      & \$23.167,22              \\ \hline
\textbf{140}                & 892-893                 & -\$442,64     & \$21.281,64              \\ \hline
\textbf{160}                & 1029-1030               & -\$364,75     & \$19.875,25              \\ \hline
\textbf{180}                & 1168-1177               & \$1.099,83    & \$19.840,19              \\ \hline
\textbf{187}                & 1211-1213               & \$2.351,19    & \$25.871,19              \\ \hline
\end{tabular}
\end{table}

\begin{table*}[!htb]
\fontsize{8}{10}\selectfont
\centering
\caption{Evaluation of Proposal with Dow Jones 30 Stocks}
\label{table:dow30_all}
\begin{tabular}{ l r r r r }
\hline
\textbf{Stock}        & \textbf{CNN-BI (2007-2012)} & \textbf{BaH (2007-2012)} & \textbf{CNN-BI (2012-2017)} & \textbf{BaH (2012-2017)} \\ \hline
\textbf{MMM}          & 7.98\%                      & 4.18\%                   & 14.40\%                     & 17.79\%                  \\ \hline
\textbf{AXP}          & -1.80\%                     & -1.68\%                  & -0.67\%                     & 8.85\%                   \\ \hline
\textbf{APPL}         & 37.46\%                     & 35.47\%                  & -1.93\%                     & 10.86\%                  \\ \hline
\textbf{BA}           & -1.62\%                     & -2.40\%                  & 10.54\%                     & 18.10\%                  \\ \hline
\textbf{CAT}          & -5.64\%                     & -0.32\%                  & -7.19\%                     & -0.32\%                  \\ \hline
\textbf{CVX}          & 16.79\%                     & 11.10\%                  & -1.32\%                     & 5.34\%                   \\ \hline
\textbf{CSCO}         & 2.69\%                      & -6.90\%                  & 14.02\%                     & 11.43\%                  \\ \hline
\textbf{KO}           & 10.65\%                     & 10.07\%                  & 1.08\%                      & 6.85\%                   \\ \hline
\textbf{DIS}          & 6.26\%                      & 2.87\%                   & 13.42\%                     & 21.07\%                  \\ \hline
\textbf{XOM}          & 0.00\%                      & 3.55\%                   & -0.90\%                     & 3.71\%                   \\ \hline
\textbf{GE}           & 0.82\%                      & -11.20\%                 & 10.54\%                     & 14.40\%                  \\ \hline
\textbf{GS}           & -18.86\%                    & -12.71\%                 & 5.02\%                      & 17.51\%                  \\ \hline
\textbf{HD}           & -0.61\%                     & 2.76\%                   & 12.73\%                     & 26.13\%                  \\ \hline
\textbf{IBM}          & 9.76\%                      & 16.29\%                  & -8.60\%                     & -0.57\%                  \\ \hline
\textbf{INTC}         & 7.87\%                      & 6.90\%                   & -0.24\%                     & 9.21\%                   \\ \hline
\textbf{JNJ}          & -0.70\%                     & 2.74\%                   & 8.93\%                      & 14.86\%                  \\ \hline
\textbf{JPM}          & 20.94\%                     & -6.05\%                  & 8.74\%                      & 20.40\%                  \\ \hline
\textbf{MCD}          & 13.34\%                     & 20.03\%                  & 4.65\%                      & 7.20\%                   \\ \hline
\textbf{MRK}          & 2.93\%                      & -0.19\%                  & 5.49\%                      & 13.13\%                  \\ \hline
\textbf{MSFT}         & -11.45\%                    & -0.80\%                  & 7.71\%                      & 18.28\%                  \\ \hline
\textbf{NKE}          & 9.23\%                      & 14.46\%                  & 3.36\%                      & 15.91\%                  \\ \hline
\textbf{PFE}          & -0.31\%                     & -0.57\%                  & 3.03\%                      & 11.52\%                  \\ \hline
\textbf{PG}           & -2.81\%                     & 2.80\%                   & 7.97\%                      & 8.72\%                   \\ \hline
\textbf{TRV}          & 2.45\%                      & 3.78\%                   & 13.46\%                     & 17.60\%                  \\ \hline
\textbf{UTX}          & 11.53\%                     & 4.51\%                   & 2.74\%                      & 7.86\%                   \\ \hline
\textbf{UNH}          & 11.06\%                     & -0.36\%                  & 11.19\%                     & 25.96\%                  \\ \hline
\textbf{VZ}           & 12.32\%                     & 6.50\%                   & 3.51\%                      & 10.84\%                  \\ \hline
\textbf{WMT}          & 1.31\%                      & 6.36\%                   & 2.05\%                      & 5.24\%                   \\ \hline
\textbf{Average}      & 7.20\%                      & 5.86\%                   & 5.84\%                      & 13.25\%                  \\ \hline
\textbf{S. Deviation} & 10.56\%                     & 9.67\%                   & 6.30\%                      & 6.97\%                   \\ \hline
\end{tabular}
\end{table*}

When we analyze (Table~\ref{table:dow30_all} and Table~\ref{table:average} ), we observe some interesting outcomes. First and foremost, the model was generally able to outperform BaH in the first case but trail in the second case (Table~\ref{table:dow30_all}). Between 2007 and 2012, the average annualized return of our proposed method (CNN-BI) was 7.20\% and the ratio of successful transactions was 52.35\%, whereas BaH average annualized return was 5.86\% indicating that the proposed method performed better when the two models are compared. Maximum annualized return in stocks was 37.46\%. However during the test period between 2012-2017, the CNN model was not able to beat BaH strategy due to the fact that the stock market (and the corresponding stocks) were in a bull market and did not suffer any significant setback for a long uninterrupted period of time. During such particular times, generally, it is not easy to outperform BaH and that was also our observation. Our bar chart image based CNN-BI model (2012-2017) had an annual return of 5.84\%, whereas BaH average annualized return was 13.25\%. Maximum annualized return in stocks in the proposed model was 14.40\%. Our strategy lost money (negative return) on 9 out of 29 times during 2007-2012, meanwhile BaH lost 10 out of 29 times during the same period. For 2012-2017 test period, these statististics were 7 out of 29 and 2 out of 29, respectively which is also another indication of a strong bull market. The overall accuracy of the model ranges between 44\% to 52\% compared to 33\% if the selection was done randomly.

\begin{table*}[!htb]
\fontsize{8}{10}\selectfont
\centering
\caption{Trade Statistics of the Proposed CNN-BI Model for Dow30}
\label{table:average}
\begin{tabular}{ l r r}
\hline
\textbf{Performance Metrics}                                 & \textbf{Average(2007-2012)} & \textbf{Average(2012-2017)} \\ \hline
\textbf{Proposed CNN-BI Strategy Annualized Return (CNN-BI)} & 7.20\%                      & 5.84\%                      \\ \hline
\textbf{Annualized Number of Transaction (AnT)}              & 52.70                       & 53.44                       \\ \hline
\textbf{Average Percent of Success (PoS)}                    & 52.35\%                     & 53.41\%                     \\ \hline
\textbf{Average Percent Profit Per Transactions (ApT)}       & 0.23\%                      & 0.18\%                      \\ \hline
\textbf{Average Transaction Length (L)}                      & 3.11                        & 3.00                        \\ \hline
\textbf{Maximum Profit Percentage in Transaction (MpT)}      & 17.30\%                     & 8.31\%                      \\ \hline
\textbf{Maximum Loss Percentage in Transaction (MlT)}        & -12.59\%                    & -8.07\%                     \\ \hline
\textbf{Maximum Capital (MaxC)}                              & \$15,767.53                 & \$14,446.25                 \\ \hline
\textbf{Minimum Capital (MinC)}                              & \$7,395.77                  & \$8,778.68                  \\ \hline
\textbf{Idle Ratio (IdleR)}                                  & 48.20\%                     & 48.88\%                     \\ \hline
\textbf{Average Percentage of Maximum Drawdown (MDD)}        & 34.92\%                     & 19.48\%                     \\ \hline
\textbf{Return Over Maximum Drawdown (RoMaD)}                & 121.31\%                    & 61.62\%                     \\ \hline
\end{tabular}
\end{table*}

The annualized profit returns in different test conditions experienced slight variances (7.20\% in 2007-2012 against 5.86\% in 2012-2017). Still, the proposed model showed some consistent operational results regardless of the underlying market (bull, bear or trendless) conditions. Table~\ref{table:average} presents the trade statistics for the proposed model in 2007-2012 and 2012-2017 periods. The results indicate that the way the model reacts to the market does not change much (i.e. annualized number of transactions, average percent success, average transaction length, etc), however, the profit performances were considerably different. As long as the annualized returns are concerned, the model performed better in 2007-2012 period when compared to 2012-2017.   As the transaction success percent indicates, the buy-sell decisions were better than random walk 19 out of 29 during the 2007-2012 period and 21 out of 29 during 2012-2017 period. The annual number of transactions ranges between 40 and 60 in both test periods, indicating the model was able to generate buy-sell transaction pairs almost once a week, regardless of the market direction. Also the average length of each transaction is 3 days during both test periods.

It is noteworthy to mention that the model sits idle almost 50\% of the time as can be seen from Table~\ref{table:average}. During trendless or bear markets, this phenomena is not much of an issue (even better in downtrending markets, since staying on cash is one of the best strategies during relentless bear markets). However, during bull markets, staying on the sidelines causes opportunity costs and lost potential profits and the overall returns are diminished if not invested in the market. That is probably the sole reason why the CNN model was not able to outperform BaH between 2012-2017, since the model did not perform any transactions and stayed on cash during 50\% of the time. Hence, if we can come up with strategies that decreases the idle rate by higher utilization, the performance results might be further improved. This can be achieved in a number of different ways (which all of them can be considered as future work): Instead of processing and trading a single stock, a basket of several stocks (preferrably uncorrelated) can be observed at any given time. Such an attempt might increase the number of annual transactions, hopefully without jeopardizing the success rate, which results in lower idle ratio and higher annual returns. Also, overall market trend might be observed and the model might be adjusted to be more aggresive during trendless or bear markets, but adapt BaH (or a comparable model) strategy during bull markets. Finally, leveraged models can also be introduced to boost the profits, since overall the success rate is above 50\% regardless of the market conditions and time periods. All these strategies might assist the model to achieve better overall performance.

\section{Conclusion }
\label{Conclusion}

In this study, we developed an out-of-the-box algorithmic trading strategy which was based on identifying Buy-Sell decisions based on triggers that are generated from a trained deep CNN model using stock bar chart images. Almost all existing strategies in literature used the stock time series data directly or indirectly, however in this study we chose a different path by using the chart images directly as 2-D images without introducing any other time series data. To best of our knowledge, this is the first attempt in literature to adapt such an unconventional approach. The results indicate that the proposed model was able to produce generally consistent outcomes and was able to beat BaH strategy depending on the market conditions. Overall the results are promising. Since, this is one of the first such attempts in this field, there is room for improvement. Increasing the number of transactions and introducing market trend as a separate feature might further enhance the performance. The model can also be integrated into an ensemble trading system with other models based on conventional methods.

\section{Acknowledgement}

This study was funded by The Scientific and Technological Research Council of Turkey (TUBITAK) under grant number 215E248.

\section*{References}



\bibliographystyle{elsarticle-num} 

\begin{thebibliography}{10}
\expandafter\ifx\csname url\endcsname\relax
  \def\url#1{\texttt{#1}}\fi
\expandafter\ifx\csname urlprefix\endcsname\relax\def\urlprefix{URL }\fi
\expandafter\ifx\csname href\endcsname\relax
  \def\href#1#2{#2} \def\path#1{#1}\fi

\bibitem{AguilarRivera2015}
R.~Aguilar-Rivera, M.~Valenzuela-Rendon, JJ.~Rodrıguez-Ortiz, {Genetic algorithms and Darwinian       approaches in financial applications: A survey}, Expert Systems with Applications 42.2, (2015) pp.   7684–-7697.
  
\bibitem{Canziani2016}
A.~Canziani, A.~Paszke, E.~Culurciello, {An analysis of deep neural network models for practical      applications} (2016) ArXiv: 1605.07678. 

\bibitem{Cavalcante2016}
R.C.~Cavalcante, R.C.~Brasileiro, V.~Souza, J.P.~Nobrega, A.L.I.~Oliveira, {Computational             intelligence and financial markets: A survey and future directions}, Expert Systems with            Applications 55, (2016) pp. 194–-211.

\bibitem{Chen2016}
J.~Chen et al., {Financial Time-Series Data Analysis Using Deep Convolutional Neural Networks},       Cloud Computing and Big Data (CCBD), 7th International Conference on. IEEE,  (2016) pp. 87–-92.

\bibitem{Chen2003}
A.~Chen, M.T.~Leung, H.~Daouk, {Application of neural networks to an emerging financial market:       forecasting and trading the Taiwan Stock Index}, Computers \& Operations Research 30.6, (2003) pp.   901–-923.

\bibitem{Ciresan2011}
D.C.~Ciresan et al., {Convolutional neural network committees for handwritten character               classification}, Document Analysis and Recognition (IC-DAR), International Conference on. IEEE,     (2011) pp. 1135–-1139.

\bibitem{Ding2015}
X.~Ding et al., {Deep Learning for Event-Driven Stock Prediction}, International Joint Conference on   Artificial Intelligence, (2015) pp. 2327–-2333.

\bibitem{Enke2013}
D.~Enke, N.~Mehdiyev, {Stock market prediction using a combination of stepwise regression analysis,   differential evolution-based fuzzy clustering, and a fuzzy inference neural network}, Intelligent   Automation \& Soft Computing 19.4, (2013) 636--648.

\bibitem{Fischer2017}
T.~Fischer, C.~Krauß, {Deep learning with long short-term memory networks for financial market        predictions}, Tech. rep. FAU Discussion Papers in Economics, (2017).

\bibitem{Goodfellow2016}
I.~Goodfellow, Y.~Bengio, A.~Courville, {Deep Learning}, MIT Press, (2016).

\bibitem{Guresen2011}
E.~Guresen, G.~Kayakutlu, T.U.~Daim, {Using artificial neural network models in stock market index    prediction}, Expert Systems with Applications 38.8, (2011) pp. 10389–-10397.

\bibitem{Karpathy2014}
A.~Karpathy et al., {Large-scale video classification with convolutional neural networks},            Proceedings of the IEEE conference on Computer Vision and Pattern Recognition, (2014) pp.      1725–-1732.

\bibitem{Kalchbrenner2014}
N.~Kalchbrenner, E.~Grefenstette, P.~Blunsom, {A convolutional neural network for modelling       sentences},  (2014) ArXiv: 1404.2188.

\bibitem{Kašćelan2015}
L.~Kašćelan, V.~Kašćelan, M.~Jovanović, {Hybrid support vector machine rule extraction method for   discovering the preferences of stock market investors: Evidence from Montenegro}, Intelligent   Automation \& Soft Computing 21.4, (2015) 503--522.

\bibitem{Krauss2017}
C.~Krauss, X.~Do, N.~Huck, {Deep neural networks, gradient-boosted trees, random forests: Statistical arbitrage on the S\&P 500}, European Journal of Operational Research 259.2, (2017) pp. 689–-702.

\bibitem{Kim2014}
Y.~Kim, {Convolutional neural networks for sentence classification}, (2014) ArXiv: 1408.5882.

\bibitem{Krizhevsky2012}
A.~Krizhevsky, I.~Sutskever, G.~Hinton, {ImageNet classification with deep convolutional neural networks}, Advances in neural information processing systems, (2012) pp. 1097–-1105.

\bibitem{Krollner2010}
B.~Krollner, B.~Vanstone, G.~Finnie, {Financial time series forecasting with machine learning techniques: A survey}, European Symposium on Artificial Neural Networks – Computational Intelligence and Machine Learning, (2010).

\bibitem{Kwon2007}
Y.K.~Kwon, B.R.~Moon, {A hybrid neurogenetic approach for stock forecasting},  IEEE Transactions on Neural Networks 18.3, (2007) pp. 851–-864.

\bibitem{Langkvist2014}
M.~Langkvist, L.~Karlsson, A.Loutfi, {A review of unsupervised feature learning and deep learning for time-series modeling}, Pattern Recognition Letters 42, (2014), pp. 11-–24.

\bibitem{Lawrence1997}
S.~Lawrence et al., {Face recognition: A convolutional neural-network approach}, IEEE transactions on neural networks 8.1., (1997) pp. 98–-113.

\bibitem{LeCun1995}
Y. LeCun et al., {Learning algorithms for classification: A comparison on handwritten digit recognition}, Neural networks: the statistical mechanics perspective, (1995) 265.

\bibitem{LeCun2015}
Y.~LeCun, Y.~Bengio, G.~Hinton, {Deep learning}, Nature 521.7553, (2015) pp. 436–-444.

\bibitem{Liao2010}
Z.~Liao, J.~Wang, {Forecasting model of global stock index by stochastic time effective neural network}, Expert Systems with Applications 37.1, (2010) pp. 834-–841.

\bibitem{Sezer2017_Artif}
O.B.~Sezer, A.M.~Ozbayoglu, E.~Dogdu, {An Artificial Neural Network-based Stock Trading System Using Technical Analysis and Big Data Framework}, ACM Proceedings of the South East Conference, (2017) pp. 223-–226.

\bibitem{Sezer2017_Deep}
O.B.~Sezer, A.M.~Ozbayoglu, E.~Dogdu, {A Deep Neural-Network based Stock Trading System based on Evolutionary Optimized Technical Analysis Parameters}, Complex Adaptive Systems Conference, 114, (2017) pp. 473--480.

\bibitem{Sezer2018}
O.B.~Sezer, A.M.~Ozbayoglu, {Algorithmic Financial Trading with Deep Convolutional Neural Networks: Time Series to Image Conversion Approach} Applied Soft Computing, 70, (2018), pp. 525--538.

\bibitem{Wang2011}
J.~Wang et al., {Forecasting stock indices with back propagation neural network}, Expert Systems with Applications 38.11, (2011) pp. 14346-–14355.



\end{thebibliography}


\end{document}